\documentclass[letterpaper]{article} % DO NOT CHANGE THIS
\usepackage[preprint]{aaai2027}  % DO NOT CHANGE THIS
\usepackage[hyphens]{url}  % DO NOT CHANGE THIS
\usepackage{graphicx} % DO NOT CHANGE THIS
\usepackage{natbib}  % DO NOT CHANGE THIS AND DO NOT ADD ANY OPTIONS TO IT
\usepackage{caption} % DO NOT CHANGE THIS AND DO NOT ADD ANY OPTIONS TO IT
\usepackage{algorithm}
\usepackage{algorithmic}

\usepackage{booktabs}
\usepackage{multirow}
\usepackage{colortbl}
\usepackage{tabularx}

\usepackage{amsmath}
\usepackage{amssymb}
\usepackage{bm}

\usepackage{subcaption}

\newcommand{\vt}{\bm{v}_\theta}
\newcommand{\zt}{\bm{z}_t}
\newcommand{\zo}{\bm{z}_0}
\newcommand{\zone}{\bm{z}_1}
\newcommand{\xact}{\bm{a}}

\newcommand{\RR}{\mathbb{R}}
\newcommand{\EE}{\mathbb{E}}

\title{Frequency-Conditioned Flow Matching for Vision-Language-Action Models}
\author{
    \normalsize \textbf{Haochen Niu$^{1}$, Shengye Dong$^{1,2}$, Hao Liu$^{1,\dagger}$, Peiwen Lin$^{1}$, Wang Chuang$^{1}$}\\[0.6em]
    \small $^{1}$AGIBOT \quad $^{2}$Xi'an Jiaotong University\\[-0.0em]
    \footnotesize $^{\dagger}$Corresponding author
}
\affiliations{}

\begin{document}

\maketitle

\begin{abstract}
Robot actions are temporally correlated trajectories whose frequency components encode motion at different scales with highly non-uniform energy distributions.
Yet Flow Matching--based vision-language-action (VLA) models typically generate actions in temporal coordinates, without explicitly modeling or systematically leveraging this frequency heterogeneity.
We introduce \emph{FreqFM}, a frequency-conditioned Flow Matching framework for VLA models.
It raises action frequency from an implicit trajectory property to an explicit conditioning dimension that spans the entire generation pipeline.
Concretely, in DCT frequency coordinates, FreqFM constructs a spectrum-matched source distribution, adaptively balances the objective across frequencies, and constrains per-frequency guidance residuals using the corresponding reference transport scales.
FreqFM integrates into existing Flow Matching action experts without changing the VLA backbone.
Across LIBERO, LIBERO-Plus, and VLA-Arena, FreqFM consistently improves performance, including a 9.3-point gain on LIBERO-Plus, and further demonstrates its effectiveness on six real-robot tasks.
\end{abstract}

\section{Introduction}

\begin{figure}[t]
\centering
\includegraphics[width=\linewidth]{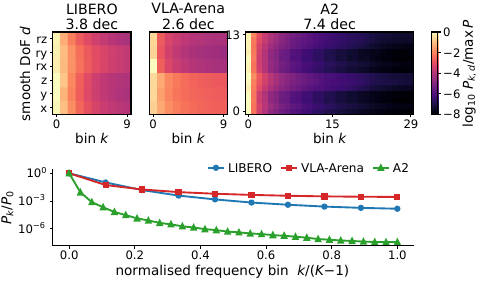}
\caption{Measured action power spectra. Top: $\log_{10}P_{k,d}$ per bin $k$ and smooth dimension $d$, normalized by the corpus maximum; headers give the dynamic range in decades. Bottom: the aggregate $P_k$, normalized by $P_0$. Horizon $H=10$ for LIBERO and VLA-Arena, $30$ for A2.}
\label{fig:spectra}
\end{figure}

Flow Matching--based vision-language-action (VLA) models generate multi-step robot actions by transporting samples from a simple source distribution to action chunks through a learned conditional velocity field \citep{lipman2023flowmatching,liu2023rectifiedflow,black2024pi0,pi05,du2026cfvla}.
Robot actions, however, are not unstructured vectors but temporally correlated trajectories.
Their frequency components describe motion at different scales: low frequencies capture global motion trends, whereas higher frequencies encode local corrections.
On our real-robot data, whose horizon is 30, energy varies by more than seven orders of magnitude across frequency bins, and more than 99.9\% is concentrated in the lowest three; on the simulation benchmarks, whose horizon is 10, the range is $3.8$ orders on LIBERO and $2.6$ on VLA-Arena (Figure~\ref{fig:spectra}).
FAST's sampling-rate study indicates that this structure affects what is learnable, not only how compactly actions can be encoded \citep{pertsch2025fast}.

Existing Flow Matching action experts typically model action generation in temporal coordinates, without explicitly modeling or systematically exploiting this frequency heterogeneity.
Neural networks can learn temporal correlations implicitly, but a frequency-selective transformation generally corresponds to dense cross-timestep coupling in temporal coordinates, while admitting a compact per-frequency parameterization in frequency coordinates.
Frequency therefore offers a structured design space for frequency-resolved modeling of action statistics.
This motivates our central question: can action frequency be elevated from an implicit trajectory property to an explicit modeling dimension for the full Flow Matching pipeline?

We answer this question with \emph{FreqFM}, a frequency-conditioned Flow Matching framework for VLA models.
FreqFM applies the discrete cosine transform (DCT) along the action horizon to expose frequency components explicitly.
For finite action chunks, the DCT yields structured coordinates with interpretable motion scales and measurable statistics, and classical results support it as a fixed basis for correlated signals \citep{ahmed1974dct}.
FreqFM then conditions three stages of Flow Matching on these coordinates:

\paragraph{Source distribution.}
FreqFM estimates a per-frequency, per-degree-of-freedom power spectrum from training actions and uses it to shape source variance.
The resulting source shares the target trajectory's spectral structure and reduces source--target covariance mismatch.
The same coupling gives each frequency a \emph{reference transport scale}: along the linear path, the target velocity at a frequency has second moment twice that frequency's own action power.

\paragraph{Training objective.}
In frequency coordinates, the relative contribution of each regression term becomes a design choice rather than a by-product of action power.
FreqFM removes that power scale by PSD normalization and then learns frequency-resolved weights through a likelihood-based multi-task objective.

\paragraph{Inference guidance.}
FreqFM measures the additional classifier-free guidance residual in units of this reference transport scale and clips it to a per-frequency budget ball.
The budget is uniform in these normalized units, so the admissible residual in raw units scales with each frequency's own reference transport scale instead of being fixed across frequencies.

FreqFM requires no modification to the VLA backbone and can be inserted into existing Flow Matching action experts.
We evaluate it on LIBERO, LIBERO-Plus, VLA-Arena, and six real-robot tasks.
FreqFM improves all three simulation benchmarks over the paired temporal Flow Matching baselines; controlled studies further examine the contribution of frequency representation and frequency conditioning across the pipeline.

\section{Related Work}

% \mathbb{E}\left[\tilde{Z}_{0},k,\mathrm{d}^2\right] = 1

% \tilde{Z}_{0,k,d} = \sqrt{P(k,d)} \, \epsilon_{k,d}

% \mathcal{L}_{\mathrm{freq}} = \sum_{k,d} \left[ \frac{1}{2} \exp(-S_{kd}) \frac{\ell_{kd}}{(P_{kd} + \epsilon)^\rho} + \frac{1}{2} S_{kd} \right]

% \hat{u}_t

% \hat{v}_\theta

% v_{\mathrm{cond}}

% v_{\mathrm{uncond}}

% M_k = \mathrm{diag}_d\left(P_{kd} + \epsilon\right)

% R_k^2 = 2\sum_{d}\frac{P_{kd}}{P_{kd}+\epsilon}

% \gamma_k = \min\left(1,\;\frac{\beta R_k}{\|w_{0,k}\|_{M_k^{-1}}}\right)

% \|w_{0,k}\|_{M_k^{-1}} \le \beta R_k

% \|w_{0,k}\|_{M_k^{-1}} > \beta R_k

% \gamma_k w_{0,k} = w_{0,k}

% \gamma_k w_{0,k}

\paragraph{Flow Matching in VLA Models.}
Diffusion Policy established diffusion over action chunks, while $\pi_0$ introduced a Flow Matching--based VLA action expert and $\pi_{0.5}$ extended it toward open-world generalization \citep{chi2024diffusionpolicy,black2024pi0,pi05}.
Recent policies make the starting point informative: some initialize from a coarse action prediction, historical actions, or a spatiotemporal warm start \citep{du2026cfvla,jia2026a2a,li2026step}, while others replace the source distribution itself with a condition-dependent prior built from action history, proprioception, or latent actions \citep{kang2026warmprior,dai2026leap,machado2026lafm}.
Classifier-free guidance (CFG) strengthens conditioning by extrapolating the conditional prediction along the conditional--unconditional difference \citep{ho2022cfg}; robotics and VLA extensions build that difference from task progress, an intentionally incoherent attention branch, or a visual-affordance baseline \citep{lu2025cfgdp,park2026acg,zhan2026rss}.
FreqFM instead matches a global, condition-independent source per frequency.

\paragraph{Frequency Structure in Robot Action Generation.}
FAST compresses action chunks into DCT-based tokens for autoregressive VLA models, while Frequency Autoregressive FreqPolicy progressively generates hierarchical frequency components with continuous tokens \citep{pertsch2025fast,zhong2025freqpolicy}.
For generative visuomotor policies, Frequency-Consistency FreqPolicy aligns DCT-domain velocities to support one-step flow generation, and FocalPolicy regularizes frequency structure across future action chunks \citep{su2025freqpolicy,he2026focalpolicy}.
FGO guides diffusion through expanding sub-frequency manifolds, and FAFM combines DCT-domain Flow Matching with derivative supervision for continuous and smooth actions \citep{wang2026fgo,guo2026fafm}.
In image diffusion, FreSca and frequency-decoupled guidance both split the CFG prediction difference into frequency bands and scale each band separately \citep{huang2025fresca,sadat2025fdg}.
Unlike methods that use frequency mainly for tokenization, consistency, smoothness, or prescribed sub-frequency traversal, FreqFM treats frequency as a conditioning dimension across the Flow Matching pipeline and budgets its frequency-wise CFG residual in units of the reference transport scale implied by the action spectrum.

\section{Frequency-Conditioned Flow Matching}

\begin{figure*}[t]
\centering
\includegraphics[width=\textwidth]{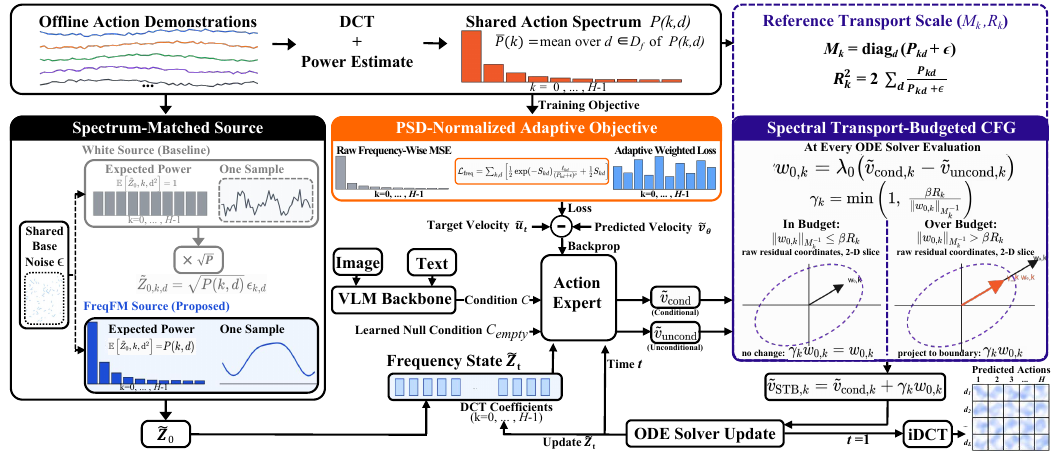}
\caption{Overview of FreqFM. Offline, a DCT of the training action chunks yields the shared action spectrum $P_{k,d}$, which conditions all three stages. \textbf{Source}: the white Gaussian source is rescaled frequency-wise to $\tilde z_{0,k,d}=\sqrt{P_{k,d}}\,\epsilon_{k,d}$, so source and target share a spectrum. \textbf{Objective}: the frequency-wise regression errors are PSD-normalized and then rebalanced by learned scales $s_{k,d}$. \textbf{Guidance}: at every ODE solver evaluation the CFG residual $\bm{w}_{0,k}$ is radially clipped into the reference-transport ball of radius $\beta R_k$ under the metric $\bm{M}_k$, leaving an in-budget residual unchanged. The VLM backbone and the action expert are unmodified.}
\label{fig:overview}
\end{figure*}

FreqFM represents smooth action dimensions in a DCT basis and conditions Flow Matching on frequency-dependent action statistics.
These coordinates make frequency directly addressable, allowing stage-specific statistics to condition source construction, objective optimization, and inference guidance within one Flow Matching formulation.
Figure~\ref{fig:overview} summarizes the three stages.

\subsection{Preliminaries}

\paragraph{Flow Matching.}
Let an action chunk be $\xact\in\RR^{H\times D}$, with horizon $H$ and action dimension $D$.
Given source $\zo\sim q_0$ and target $\zone=\xact\sim q_1$, Flow Matching \citep{lipman2023flowmatching,liu2023rectifiedflow} learns a conditional velocity field $\vt(\zt,t,c)$ along the linear path
\begin{equation}
    \zt=(1-t)\zo+t\zone,
    \qquad
    \bm{u}_t=\zone-\zo,
\end{equation}
by minimizing $\EE\|\vt(\zt,t,c)-\bm{u}_t\|_2^2$.
At inference, the learned ODE transports a source sample to an action chunk.

\paragraph{Frequency coordinates.}
Let $\bm{C}\in\RR^{H\times H}$ denote the orthonormal DCT-II matrix \citep{ahmed1974dct}.
FreqFM transforms the horizon axis as
\begin{equation}
    \tilde{\xact}=\bm{C}\xact,
    \qquad
    \xact=\bm{C}^{\top}\tilde{\xact}.
    \label{eq:dct}
\end{equation}
The empirical covariance becomes $\bm{\Sigma}_f=\bm{C}\bm{\Sigma}_t\bm{C}^{\top}$.
For action chunks with strong temporal correlation, we use the DCT-II as a fixed, data-independent frequency basis; classical comparisons show that the DCT can approach the coding performance of the data-dependent Karhunen--Lo\`eve transform on correlated sources \citep{ahmed1974dct}.
Strong spectral concentration creates large per-frequency scale differences, making diagonal scale correction a primary target.
FreqFM therefore uses per-frequency action power as a tractable first approximation to the full frequency-domain second-moment structure.

Flow Matching operates on DCT coefficients for smooth dimensions.
The final action chunk is recovered by inverse DCT after ODE integration.

\subsection{Spectrum-Matched Source}

The source distribution determines both the starting point of each training reference path and the initial state of the inference ODE.
Consider the standard centered isotropic source $q_0=\mathcal{N}(\bm{0},\bm{I})$.
Because the DCT is orthonormal, this source remains flat in frequency coordinates, with $\EE[\tilde z_{0,k,d}^2]=1$ for every frequency.
Robot actions instead have a highly non-uniform spectrum.
The standard linear path therefore connects endpoints with different relative spectral scales, and its second-moment spectrum changes shape over transport time.

FreqFM retains a centered, condition-independent Gaussian source and shapes only its frequency-wise scale using training actions.
We define the RMS coefficient scale and corresponding per-frequency power for frequency $k$ and smooth action dimension $d$ as
\begin{equation}
    A_{k,d}
    =
    \sqrt{\EE_{\xact\sim\mathcal{D}}[\tilde a_{k,d}^{2}]},
    \qquad
    P_{k,d}=A_{k,d}^{2}.
    \label{eq:spectral_amplitude}
\end{equation}
FreqFM samples the source in frequency coordinates as
\begin{equation}
    \tilde z_{0,k,d}
    =
    A_{k,d}\epsilon_{k,d}
    =
    \sqrt{P_{k,d}}\epsilon_{k,d},
    \qquad
    \epsilon_{k,d}\sim\mathcal{N}(0,1).
    \label{eq:colored_source}
\end{equation}
Thus, $q_0=\mathcal{N}(\bm{0},\operatorname{diag}(\bm{P}))$ and $\EE[\tilde z_{0,k,d}^{2}]=P_{k,d}$.

The linear Flow Matching path gives a direct interpretation.
For independently sampled source and target, the centered source yields $\EE[\tilde z_{0,k,d}\tilde a_{k,d}]=0$, so the spectrum-matched path satisfies
\begin{equation}
    \EE[\tilde z_{t,k,d}^{2}]
    =
    (1-t)^2P_{k,d}+t^2P_{k,d}
    =
    \left[(1-t)^2+t^2\right]P_{k,d}.
    \label{eq:matched_path_spectrum}
\end{equation}
All frequencies share the same time factor, preserving the relative second-moment spectrum along the reference path.
By contrast, a white source gives $\EE[\tilde z_{t,k,d}^{2}]=(1-t)^2+t^2P_{k,d}$, whose relative spectrum varies with $t$.
Spectrum matching therefore preserves the relative second-moment spectrum along the reference path, avoiding the frequency-dependent rescaling induced by a white source.

The quantity matched here is the second raw moment $P_{k,d}=\EE[\tilde a_{k,d}^2]$ rather than the coefficient variance; matching the variance would leave $\EE[\tilde z_{t,k,d}^2]=(1-t)^2\sigma_{k,d}^2+t^2P_{k,d}$, whose shape again depends on $t$.
The same choice supports a complementary marginal reading under a diagonal-Gaussian approximation.
Let the target marginal be approximated by $\mathcal{N}(\bm{\mu},\operatorname{diag}(\bm{\sigma}^2))$.
For the centered source in Eq.~\eqref{eq:colored_source}, the squared Wasserstein-2 distance becomes
\begin{equation}
    W_2^2
    \approx
    \lVert\bm{\mu}\rVert_2^2
    +
    \sum_{k,d}
    \left(\sqrt{P_{k,d}}-\sigma_{k,d}\right)^2.
    \label{eq:w2_diag}
\end{equation}
Because $P_{k,d}=\sigma_{k,d}^2+\mu_{k,d}^2$, the second term is fourth order in $\mu_{k,d}/\sigma_{k,d}$, so raw-moment matching also approximates covariance matching when the empirical coefficient means are small relative to their RMS coefficient scales; a centered source does not model the remaining $\lVert\bm{\mu}\rVert_2^2$.
We use this Gaussian calculation only as an auxiliary second-order interpretation; the pathwise result in Eq.~\eqref{eq:matched_path_spectrum} does not require a zero-mean target or a Gaussian target distribution.

\subsection{PSD-Normalized Adaptive Objective}

Frequency coordinates make individual regression terms directly addressable, allowing FreqFM to adapt their relative contributions during optimization.
We formulate this adaptation using likelihood-based multi-task weighting.
Directly learning weights from raw frequency-domain losses, however, would conflate action-power scale with variation in model residuals.
For the same relative prediction error, a component with power $P_{k,d}$ produces a squared error proportional to $P_{k,d}$; summing raw MSE therefore lets high-energy low-frequency components dominate the loss and gradients.
FreqFM consequently normalizes coefficient-wise losses using the training action spectrum before learning their adaptive weights.
Let
\begin{equation}
    \ell_{k,d}
    =
    \left(\tilde{\bm{v}}_\theta(\tilde{\bm{z}}_t,t,c)_{k,d}-\tilde u_{t,k,d}\right)^2
\end{equation}
denote the velocity-regression error for modeled frequency $k$ and smooth action dimension $d$.
Defining the power-normalized relative residual
$\delta_{k,d}=(\tilde{\bm{v}}_\theta(\tilde{\bm{z}}_t,t,c)_{k,d}-\tilde u_{t,k,d})/(P_{k,d}+\varepsilon)^{1/2}$ gives
$\ell_{k,d}=(P_{k,d}+\varepsilon)\delta_{k,d}^2$.
The raw objective therefore weights comparable relative errors in direct proportion to frequency power.
We first normalize its scale using the training spectrum,
\begin{equation}
    \bar\ell_{k,d}
    =
    \frac{\ell_{k,d}}{(P_{k,d}+\varepsilon)^{\rho}},
    \qquad
    \bar r_{k,d}
    =
    \frac{\tilde{\bm{v}}_\theta(\tilde{\bm{z}}_t,t,c)_{k,d}-\tilde u_{t,k,d}}
    {(P_{k,d}+\varepsilon)^{\rho/2}},
    \label{eq:psd_normalized_error}
\end{equation}
where $\bar\ell_{k,d}=\bar r_{k,d}^2=(P_{k,d}+\varepsilon)^{1-\rho}\delta_{k,d}^2$ and $\rho$ controls the normalization strength.
When $\rho=1$, the normalization removes the power scale completely; we use a tempered $\rho=0.25$ to reduce low-frequency dominance without excessively amplifying very small high-frequency coefficients.
The normalized losses therefore provide a more scale-balanced starting point for joint optimization across frequency bins.

For the normalized coefficient-wise regression terms, we adopt the likelihood-based multi-task weighting formulation of \citet{kendall2018multitask}.
We assume $\bar r_{k,d}\sim\mathcal{N}(0,\tau_{k,d}^2)$ and learn one log scale $s_{k,d}=\log\tau_{k,d}^2$ for each $(k,d)$ regression term.
The negative log-likelihood, up to constants, is
\begin{equation}
    \mathcal{L}_{\mathrm{freq}}
    =
    \sum_{k,d}
    \left[
        \frac{1}{2}\exp(-s_{k,d})\bar\ell_{k,d}
        +\frac{1}{2}s_{k,d}
    \right].
    \label{eq:adaptive_loss}
\end{equation}
The benefit of normalization appears directly at the start of optimization.
With $s_{k,d}=0$, the gradients for the velocity model and adaptive scales are
\begin{equation}
    \begin{aligned}
        \nabla_{\theta}\mathcal{L}_{\mathrm{freq}}\big|_{s=0}
        &=\frac{1}{2}\sum_{k,d}\nabla_{\theta}\bar\ell_{k,d}, \\
        \frac{\partial\mathcal{L}_{k,d}}{\partial s_{k,d}}\bigg|_{s=0}
        &=\frac{1}{2}\left(1-\bar\ell_{k,d}\right).
    \end{aligned}
    \label{eq:adaptive_loss_initial_gradients}
\end{equation}
PSD normalization thus reduces the known power-induced dynamic range seen by both updates before the learned scales have adapted.
During joint finite-step training, this preconditioning limits early domination by high-energy components and reduces the range that the adaptive scales must learn; the log-scale term prevents a degenerate unbounded scale.
At an unconstrained inner optimum of the expected objective for a fixed velocity field, $s_{k,d}^{*}=\log\EE[\bar\ell_{k,d}]$, and substituting back gives $\mathcal{L}_{\mathrm{freq}}^{*}=\frac{1}{2}\sum_{k,d}\log\EE[\ell_{k,d}]+\mathrm{const}$: the adaptive scales turn the objective into a log of the expected per-coefficient loss, which is invariant to any fixed reweighting.
Normalization therefore does not change the fully optimized stationary weighting by itself.
Its practical effect comes from the parameterization of the joint optimization, including zero initialization and finite training.
% , and the bounded effective-weight ratio used for stability.

\subsection{Spectral Transport-Budgeted Classifier-Free Guidance}

Standard CFG applies a single extrapolation coefficient to every action frequency.
Action power, however, varies by orders of magnitude across frequency bins, so a single coefficient produces frequency-wise extrapolation residuals whose magnitudes bear no fixed relation to the reference transport scale of each frequency.
We therefore constrain only the additional CFG residual, using the reference transport scale induced by the spectrum-matched source--target coupling to define frequency-normalized units.

\paragraph{Reference transport scale.}
Let $\mathcal{D}_f$ denote the smooth action dimensions represented by DCT coefficients.
The target coefficient and the spectrum-matched source of Eq.~\eqref{eq:colored_source} satisfy $\EE[\tilde z_{1,k,d}^2]=\EE[\tilde z_{0,k,d}^2]=P_{k,d}$.
Because the source is centered and sampled independently of the target, $\EE[\tilde z_{0,k,d}\tilde z_{1,k,d}]=0$, and the linear-path target velocity $\tilde u_{t,k,d}=\tilde z_{1,k,d}-\tilde z_{0,k,d}$ has second moment
\begin{equation}
    \EE[\tilde u_{t,k,d}^2]
    =
    \EE[\tilde z_{1,k,d}^2]+\EE[\tilde z_{0,k,d}^2]
    -2\EE[\tilde z_{0,k,d}\tilde z_{1,k,d}]
    =
    2P_{k,d}.
    \label{eq:reference_transport}
\end{equation}
This identity fixes a frequency-resolved reference scale for transport velocity, which STB-CFG uses as the unit in which the guidance residual is measured.
For a numerical stabilizer $\varepsilon>0$, we define the frequency-wise metric and reference radius
\begin{equation}
    \begin{aligned}
        \bm{M}_k&=\operatorname{diag}_{d\in\mathcal{D}_f}(P_{k,d}+\varepsilon), \\
        R_k^2
        &=
        \EE\left[\lVert\tilde{\bm{u}}_{t,k}\rVert_{\bm{M}_k^{-1}}^2\right]
        =
        2\sum_{d\in\mathcal{D}_f}
        \frac{P_{k,d}}{P_{k,d}+\varepsilon},
    \end{aligned}
    \label{eq:reference_radius}
\end{equation}
where $\lVert\bm{x}\rVert_{\bm{M}_k^{-1}}^2=\bm{x}^{\top}\bm{M}_k^{-1}\bm{x}$.
Without $\varepsilon$, this metric assigns reference second moment $2$ to every coordinate with positive power, so $R_k^2=2|\mathcal{D}_f|$ takes the same value at every frequency.
The budget is therefore uniform in reference-transport units, and all of its frequency dependence is carried by the metric $\bm{M}_k$: in raw units the admissible residual scales with the RMS coefficient scale $A_{k,d}=\sqrt{P_{k,d}}$, which spans nearly two orders of magnitude across frequency bins on LIBERO and close to four orders on our real-robot data.
The stabilizer prevents very small-power coordinates from receiving excessive inverse weights; we keep $\varepsilon$ well below $\min_{k,d}P_{k,d}$ so that the induced variation in $R_k$ remains negligible.
Unlike the tempered exponent $\rho$ in the training objective, Eq.~\eqref{eq:reference_radius} defines inference-time reference units only.

\paragraph{Frequency-wise projection of the CFG residual.}
At solver time $t$, let $\tilde{\bm{v}}_{\mathrm{cond},k}(t)$ and $\tilde{\bm{v}}_{\mathrm{uncond},k}(t)$ collect the conditional and unconditional velocity predictions for frequency $k$ over $\mathcal{D}_f$, where the unconditional branch uses the learned null condition.
We write the conditional residual and the CFG extrapolation as
\begin{equation}
    \begin{aligned}
        \bm{\Delta}_k(t)
        &=
        \tilde{\bm{v}}_{\mathrm{cond},k}(t)-\tilde{\bm{v}}_{\mathrm{uncond},k}(t), \\
        \bm{w}_{0,k}(t)&=\lambda_0\bm{\Delta}_k(t),
    \end{aligned}
    \label{eq:cfg_residual}
\end{equation}
where $\lambda_0\ge0$ is selected via a validation sweep and then frozen; if CFG is written as $\tilde{\bm{v}}_{\mathrm{uncond}}+s_0(\tilde{\bm{v}}_{\mathrm{cond}}-\tilde{\bm{v}}_{\mathrm{uncond}})$, then $\lambda_0=s_0-1$.
For every frequency, STB-CFG projects only the added extrapolation residual into a reference-transport ball,
\begin{equation}
    \bm{w}_k^{*}(t)
    =
    \operatorname*{arg\,min}_{\bm{w}^{\top}\bm{M}_k^{-1}\bm{w}\le\beta^2R_k^2}
    \frac{1}{2}
    \lVert\bm{w}-\bm{w}_{0,k}(t)\rVert_{\bm{M}_k^{-1}}^2,
    \label{eq:stb_projection}
\end{equation}
where the dimensionless budget multiplier $\beta\ge0$ is selected on validation after $\lambda_0$ has been frozen, and one shared $\beta$ is used across all frequencies and solver steps.
Writing $b_k(t)=\lVert\bm{w}_{0,k}(t)\rVert_{\bm{M}_k^{-1}}$, the exact projection admits the closed form
\begin{equation}
    \begin{aligned}
        \gamma_k(t)
        &=
        \begin{cases}
            1, & b_k(t)\le\beta R_k,\\[2pt]
            \beta R_k/b_k(t), & b_k(t)>\beta R_k,
        \end{cases} \\[2pt]
        \bm{w}_k^{*}(t)&=\gamma_k(t)\bm{w}_{0,k}(t),
    \end{aligned}
    \label{eq:stb_scale}
\end{equation}
so the guided velocity is
\begin{equation}
    \tilde{\bm{v}}_{\mathrm{STB},k}(t)
    =
    \tilde{\bm{v}}_{\mathrm{cond},k}(t)+\bm{w}_k^{*}(t).
    \label{eq:guidance_scale}
\end{equation}
STB-CFG leaves an in-budget CFG residual unchanged and radially clips an out-of-budget residual at the boundary.
The limit $\beta\rightarrow\infty$ recovers standard CFG, while $\beta=0$ recovers conditional inference on the projected smooth dimensions.
The scale $\gamma_k(t)$ is recomputed for every sample and solver evaluation.
Conditional and unconditional velocity estimates share the same action expert, and during training we replace the condition prefix with a learned null embedding with probability $p_{\mathrm{drop}}$.

\section{Experiments}

\subsection{Experimental Setup}

\paragraph{Backbones and implementation.}
We instantiate FreqFM on both $\pi_0$ \citep{black2024pi0} and $\pi_{0.5}$ \citep{pi05} using the LeRobot codebase \citep{cadene2026lerobot}.
For each backbone, its temporal Flow Matching counterpart uses the same architecture, data, optimizer, and training budget; only the Flow Matching formulation differs.
Each FreqFM model and its paired temporal Flow Matching baseline are trained for 20k optimization steps on two NVIDIA A800 GPUs with a global batch size of 64.
Unless stated otherwise, we follow the official backbone configurations and benchmark evaluation protocols.
We use a PSD-normalization exponent $\rho=0.25$ and set the condition-dropout probability to $p_{\mathrm{drop}}=0.05$.
Frequency statistics are estimated only from training actions.
LIBERO-Plus is evaluated zero-shot from the LIBERO-trained checkpoints, which also reuse the frequency statistics estimated on LIBERO training data.
Further details are provided in the supplementary material.

\paragraph{Benchmarks and metrics.}
LIBERO comprises four suites of ten manipulation tasks each, evaluated over 50 episodes per task.
LIBERO-Plus applies seven perturbation families to the four LIBERO suites, producing more than ten thousand perturbed task instances, each evaluated for one episode.
VLA-Arena contains eleven tasks grouped into four categories. Each task provides five instances at each difficulty level (L0--L2), except Long-Horizon, which provides ten L0 instances; each instance is evaluated over 10 episodes.
We report success rate (SR); on VLA-Arena we average over the instances within a level, then over the three levels, and finally over the eleven tasks to obtain Total.

\subsection{Simulation Results}

\begin{table}[t]
\centering
{
\small
\setlength{\tabcolsep}{4pt}
\begin{tabular}{lccccc}
\toprule
\textbf{Method} & \textbf{Spatial} & \textbf{Object} & \textbf{Goal} & \textbf{Long} & \textbf{Avg.} \\
\midrule
UniVLA & 96.5 & 96.8 & 95.6 & 92.0 & 95.2 \\
OpenVLA-OFT & 97.6 & 98.4 & \textbf{97.9} & 94.5 & 97.1 \\
$\pi_0$-Fast & 96.4 & 96.8 & 88.6 & 60.2 & 85.5 \\
GR00T N1 & 94.4 & 97.6 & 93.0 & 90.6 & 93.9 \\
VLA-JEPA & 94.8 & \textbf{99.6} & 95.8 & 94.0 & 96.1 \\
FreqPolicy & 97.0 & 98.6 & 96.0 & 87.6 & 94.8 \\
\midrule
$\pi_0^\dagger$ & 93.2 & 98.4 & 96.4 & 93.6 & 95.4 \\
FreqFM ($\pi_0$) & 95.8 & 97.8 & 95.6 & 95.2 & 96.1 \\
\rowcolor[gray]{0.94}
$\Delta$ & +2.6 & -0.6 & -0.8 & +1.6 & +0.7 \\
\midrule
$\pi_{0.5}^\dagger$ & 95.2 & \textbf{99.6} & 97.2 & 94.6 & 96.7 \\
FreqFM ($\pi_{0.5}$) & \textbf{98.6} & 99.4 & 97.6 & \textbf{95.8} & \textbf{97.9} \\
\rowcolor[gray]{0.94}
$\Delta$ & +3.4 & -0.2 & +0.4 & +1.2 & +1.2 \\
\bottomrule
\end{tabular}
}
\caption{Success rate on the four LIBERO suites. Best per column in bold. Each $\Delta$ is the absolute percentage-point change over the paired temporal Flow Matching baseline. $\dagger$ marks our reproduction, trained and evaluated under the same backbone architecture, data, optimizer, training budget, and evaluation protocol as its paired FreqFM row.}
\label{tab:libero}
\end{table}

\begin{table*}[t]
\centering
{
\small
\setlength{\tabcolsep}{4pt}
\begin{tabularx}{\textwidth}{l*{8}{>{\centering\arraybackslash}X}}
\toprule
\textbf{Method}
& \textbf{Camera}
& \textbf{Robot}
& \textbf{Language}
& \textbf{Light}
& \textbf{Background}
& \textbf{Noise}
& \textbf{Layout}
& \textbf{Total} \\
\midrule
OpenVLA-OFT & 55.6 & 21.7 & 81.0 & 92.7 & 91.0 & \textbf{78.6} & 68.7 & 67.9 \\
RIPT-VLA & 55.2 & 31.2 & 77.6 & 88.4 & \textbf{91.6} & 73.5 & 74.2 & 68.4 \\
UniVLA & 1.8 & 46.2 & 69.6 & 69.0 & 81.0 & 21.2 & 31.9 & 42.9 \\
% $\pi_0$ & 13.8 & 6.0 & 58.8 & 85.0 & 81.4 & 79.0 & 68.9 & 53.6 \\
$\pi_0$-Fast & \textbf{65.1} & 21.6 & 61.0 & 73.2 & 73.2 & 74.4 & 68.8 & 61.6 \\
VLA-JEPA & 40.3 & 55.7 & 72.9 & 88.2 & 70.5 & 38.2 & 74.6 & 62.9 \\
\midrule
$\pi_0^\dagger$ & 45.5 & 31.5 & 61.9 & 95.0 & 87.1 & 56.9 & 77.8 & 62.7 \\
FreqFM ($\pi_0$) & 53.4 & 41.1 & 67.3 & 92.2 & 82.9 & 71.5 & 81.6 & 68.2 \\
\rowcolor[gray]{0.94}
$\Delta$ & +7.9 & +9.6 & +5.4 & -2.8 & -4.2 & +14.6 & +3.8 & +5.5 \\
\midrule
$\pi_{0.5}^\dagger$ & 41.7 & 62.4 & 76.7 & 96.4 & 88.0 & 37.8 & 79.2 & 66.5 \\
FreqFM ($\pi_{0.5}$) & 58.8 & \textbf{77.2} & \textbf{81.5} & \textbf{97.0} & 89.1 & 53.5 & \textbf{84.4} & \textbf{75.8} \\
\rowcolor[gray]{0.94}
$\Delta$ & +17.1 & +14.8 & +4.8 & +0.6 & +1.1 & +15.7 & +5.2 & +9.3 \\
\bottomrule
\end{tabularx}
}
\caption{Success rate under the seven LIBERO-Plus perturbation types, averaged over the four underlying LIBERO suites. Bolding and $\dagger$ follow Table~\ref{tab:libero}.}
\label{tab:libero_plus}
\end{table*}

\begin{table*}[t]
\centering
{
\small
\setlength{\tabcolsep}{4pt}
\begin{tabularx}{\textwidth}{l*{12}{>{\centering\arraybackslash}X}}
\toprule
\multirow{2}{*}{\textbf{Method}}
& \multicolumn{5}{c}{\textbf{Safety}}
& \multicolumn{2}{c}{\textbf{Distractor}}
& \multicolumn{3}{c}{\textbf{Extrapolation}}
& \multirow{2}{*}{\textbf{LH}}
& \multirow{2}{*}{\textbf{Total}} \\
\cmidrule(lr){2-6}\cmidrule(lr){7-8}\cmidrule(lr){9-11}
& \textbf{SO} & \textbf{CG} & \textbf{HA} & \textbf{SP} & \textbf{DO}
& \textbf{SD} & \textbf{DD}
& \textbf{PC} & \textbf{TW} & \textbf{UO}
& & \\
\midrule
% $\pi_0$ & 68.0 & 28.7 & 26.7 & 70.7 & 58.0 & 43.3 & 55.3 & 26.7 & 26.0 & 55.3 & 32.3 & 44.6 \\
$\pi_0$-Fast & 73.3 & 37.3 & 26.0 & 78.0 & 55.3 & 46.7 & 70.7 & 24.0 & 30.0 & 20.0 & 27.3 & 44.4 \\
OpenVLA-OFT & 46.7 & 36.7 & 18.7 & 65.3 & 48.7 & 40.0 & 64.7 & 26.7 & 24.7 & 40.0 & 26.7 & 39.9 \\
OpenVLA & 40.0 & 40.0 & 14.0 & 66.7 & 48.7 & 33.3 & 52.7 & 24.0 & 39.3 & 46.7 & 26.7 & 39.3 \\
UniVLA & 48.0 & \textbf{46.7} & 28.7 & 73.3 & 30.7 & 37.3 & 45.3 & 18.0 & 33.3 & 42.0 & 22.0 & 38.7 \\
Qwen3-VL-Fast & 36.7 & 42.7 & 35.3 & 58.0 & 42.0 & 39.3 & 51.3 & 17.3 & 25.3 & 39.3 & 33.3 & 38.2 \\
Motus & 52.3 & 34.0 & 27.7 & 60.3 & 34.7 & 32.3 & 55.3 & 4.3 & 11.3 & 42.3 & 24.0 & 34.4 \\
\midrule
$\pi_0^\dagger$ & 77.3 & 42.7 & 46.0 & 84.7 & 66.7 & 52.7 & 67.3 & 28.7 & 34.0 & \textbf{70.0} & 32.0 & 54.7 \\
FreqFM ($\pi_0$) & 77.3 & 43.3 & 53.3 & \textbf{93.3} & \textbf{80.0} & 56.7 & 66.7 & \textbf{30.0} & 26.7 & 66.7 & 36.7 & 57.3 \\
\rowcolor[gray]{0.94}
$\Delta$ & 0.0 & +0.6 & +7.3 & +8.6 & +13.3 & +4.0 & -0.6 & +1.3 & -7.3 & -3.3 & +4.7 & +2.6 \\
\midrule
$\pi_{0.5}^\dagger$ & 96.0 & 40.7 & 73.3 & 88.0 & 62.7 & 50.7 & 92.0 & 28.7 & \textbf{59.3} & 66.7 & 45.7 & 64.0 \\
FreqFM ($\pi_{0.5}$) & \textbf{99.3} & \textbf{46.7} & \textbf{84.7} & 92.0 & 78.7 & \textbf{58.7} & \textbf{93.3} & 25.3 & 55.3 & 63.3 & \textbf{53.3} & \textbf{68.2} \\
\rowcolor[gray]{0.94}
$\Delta$ & +3.3 & +6.0 & +11.4 & +4.0 & +16.0 & +8.0 & +1.3 & -3.4 & -4.0 & -3.4 & +7.6 & +4.2 \\
\bottomrule
\end{tabularx}
}
\caption{Success rate on VLA-Arena. The eleven tasks are abbreviated by their initials and grouped into four categories. \emph{Safety}: SO = Static Obstacles, CG = Cautious Grasp, HA = Hazard Avoidance, SP = State Preservation, DO = Dynamic Obstacles. \emph{Distractor}: SD = Static Distractors, DD = Dynamic Distractors. \emph{Extrapolation}: PC = Preposition Combinations, TW = Task Workflows, UO = Unseen Objects. \emph{Long-Horizon}: LH = Long Horizon. Each entry averages Levels 0, 1, and 2, and Total averages over all eleven tasks. Bolding and $\dagger$ follow Table~\ref{tab:libero}.}
\label{tab:vla_arena}
\end{table*}

We compare FreqFM with its paired temporal Flow Matching baseline on both backbones, and with prior methods, on the three simulation benchmarks \citep{liu2024libero,fei2025liberoplus,arena2026}: UniVLA \citep{bu2025univla}, OpenVLA-OFT \citep{kim2025openvlaoft}, $\pi_0$-Fast \citep{pertsch2025fast}, GR00T N1 \citep{bjorck2025gr00tn1}, VLA-JEPA \citep{sun2026vlajepa}, FreqPolicy \citep{su2025freqpolicy}, RIPT-VLA \citep{tan2025riptvla}, OpenVLA \citep{kim2024openvla}, Qwen3-VL-Fast \citep{starvla2026}, and Motus \citep{bi2026motus}.

\paragraph{LIBERO.}
In Table~\ref{tab:libero} all methods are near saturation, and the gains are correspondingly small ($+0.7$ and $+1.2$ points).

\paragraph{LIBERO-Plus.}
The largest gains appear here (Table~\ref{tab:libero_plus}; $+5.5$ and $+9.3$ points).
On both backbones the three perturbations with the largest gain, Camera, Robot, and Noise, are exactly the three on which the temporal baseline is weakest, and the gain correlates negatively with the baseline success rate ($-0.81$ on $\pi_0$ and $-0.95$ on $\pi_{0.5}$).
Light and Background, the two perturbations the baseline already handles above $87$ points, are also the only two on which $\pi_0$ regresses ($-2.8$ and $-4.2$).

\paragraph{VLA-Arena.}
In Table~\ref{tab:vla_arena} the total improves by $+2.6$ and $+4.2$ points, while the Extrapolation category regresses on both backbones.
FreqFM estimates its source spectrum and transport budget once on the training actions and then keeps both fixed, so tasks whose action statistics depart from the training distribution are conditioned on a mismatched spectral prior.

The gain is consistently larger on $\pi_{0.5}$ than on $\pi_0$ on all three benchmarks, even though $\pi_{0.5}$ is the stronger baseline of the two.
We attribute this to pre-training: $\pi_{0.5}$ is pre-trained with FAST action tokens alongside its Flow Matching action expert, and FAST tokenizes actions through the DCT, so its backbone has already been exposed to a frequency-domain action representation.

\subsection{Real-World Evaluation}

\begin{figure}[t]
\centering
\includegraphics[width=\linewidth]{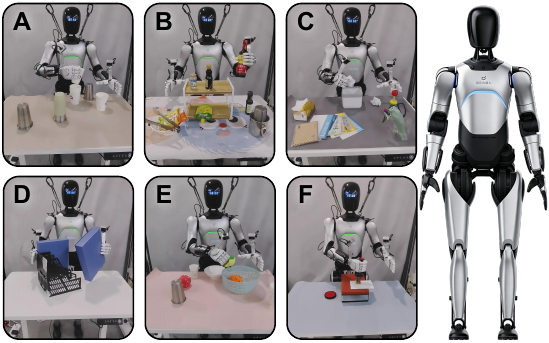}
\caption{Real-robot task suite and the AgiBot Expedition A2 platform (right). (A) Cup Stacking, (B) Kitchen Tidy, (C) Table Cleanup, (D) Folder Filing, (E) Toy Storage, (F) Stamp \& Handover.}
\label{fig:realworld}
\end{figure}

We evaluate FreqFM on an AgiBot Expedition A2 humanoid across six tasks with different actuation scopes (Figure~\ref{fig:realworld}).
Policies receive RGB observations from one chest-mounted and two wrist-mounted cameras and control task-specific arm and waist actions.
The single-arm tasks require the left arm to place a folder into a file organizer and the right arm to place toys from a table into a storage bin.
The bimanual tasks comprise placing condiments onto a spice rack, dropping crumpled paper into a desktop bin, and adding further cups to an existing stack of paper cups.
The final task coordinates both arms and the waist to stamp a postcard, turn, and hand it to a person.
This suite covers single-arm, bimanual, and whole-upper-body control within one hardware platform.
Every task is evaluated over 30 trials.
Every task decomposes into several subtask stages, so we report a subtask success rate alongside the overall success rate. The stage definitions are given in the supplementary material.

Table~\ref{tab:real_robot} reports the results.
On Toy Storage, the easiest task, both policies are already near ceiling and reach the same success rate.
The gain concentrates on the tasks whose stages require fine positional alignment: stacking the cups, inserting the folder into a slot of the organizer, and pinching the stamped postcard off the table gain $+23.3$, $+13.4$, and $+13.3$ points, against $+6.7$ and $+3.3$ on the two tasks whose placements have loose tolerances.
Such stages depend on small terminal corrections, which occupy the low-energy high-frequency end of the measured action spectrum (Figure~\ref{fig:spectra}) that a raw MSE objective weights least.

\begin{table}[t]
\centering
{
\small
\setlength{\tabcolsep}{4pt}
\begin{tabular}{@{}l c cc cc@{}}
\toprule
\multirow{2}{*}{\textbf{Task}} & \multirow{2}{*}{\textbf{Stages}} & \multicolumn{2}{c}{$\bm{\pi}_{\bm{0.5}}$} & \multicolumn{2}{c}{\textbf{FreqFM}} \\
\cmidrule(lr){3-4}\cmidrule(lr){5-6}
& & \textbf{Sub-SR} & \textbf{SR} & \textbf{Sub-SR} & \textbf{SR} \\
\midrule
Folder Filing & 3 & 64.4 & 53.3 & 77.8 & 66.7 \\
Toy Storage & 4 & 91.7 & 90.0 & 94.2 & 90.0 \\
\addlinespace[1pt]
Kitchen Tidy & 4 & 69.2 & 56.7 & 72.5 & 60.0 \\
Table Cleanup & 4 & 70.0 & 60.0 & 78.3 & 66.7 \\
Cup Stacking & 4 & 57.5 & 30.0 & 71.7 & 53.3 \\
\addlinespace[1pt]
Stamp \& Handover & 5 & 70.7 & 56.7 & 79.3 & 70.0 \\
\midrule
\textbf{Average} & & 70.6 & 57.8 & 79.0 & 67.8 \\
\bottomrule
\end{tabular}
}
\caption{Real-world success rate (SR) and subtask success rate (Sub-SR) on the AgiBot Expedition A2, over 24 subtask stages in total. Stages is the number of task-defined subtask stages. SR is the fraction of trials that complete the full task; Sub-SR credits partial progress by averaging the fraction of stages completed per trial. Averages are over the six tasks.}
\label{tab:real_robot}
\end{table}

\subsection{Ablations and Analysis}

\paragraph{Ablation studies.}
We first compare FreqFM with a temporal control that instantiates time-domain counterparts of the three stages, and then remove one component at a time from the complete frequency-conditioned framework.
The temporal control does improve on the temporal Flow Matching baseline ($+0.3$ and $+2.5$ points), but by far less than FreqFM ($+1.2$ and $+9.3$). 
The three mechanisms are not sufficient on their own, and their benefit depends on coordinates in which the frequency structure is explicit.
Among the removals, the matched source matters most, costing $4.2$ points on LIBERO-Plus.
LIBERO is saturated and separates the variants by at most $0.7$ point, so we read the component ordering from LIBERO-Plus.
The transport budget produces the smallest success-rate change ($1.9$ points), but its effect is on trajectory quality rather than on task completion.
At a comparable success rate on LIBERO, budgeting the guidance residual cuts the mean absolute jerk of the executed arm trajectories by $56\%$, from $0.442$ under standard CFG to $0.193$, where jerk is the time-averaged $\ell_2$ norm of the third-order difference of the commanded joint trajectory.
The budget is uniform in reference-transport units, so in raw units it is tightest at the low-power high-frequency bins, which are exactly the bins in which unbudgeted extrapolation injects the fastest motion.

\begin{table}[t]
\centering
{
\small
\setlength{\tabcolsep}{4pt}
\begin{tabular}{@{}lcc@{}}
\toprule
\textbf{Variant} & \textbf{LIBERO} & \textbf{LIBERO-Plus} \\
\midrule
$\pi_{0.5}$ (Temporal FM) & 96.7 & 66.5 \\
\quad + Time-Domain Components & 97.0 & 69.0 \\
\midrule
FreqFM & 97.9 & 75.8 \\
\quad w/o Matched Source & 97.2 & 71.6 \\
\quad w/o Adaptive Objective & 97.9 & 72.2 \\
\quad w/o Transport Budget & 97.8 & 73.9 \\
\bottomrule
\end{tabular}
}
\caption{Ablation studies on LIBERO and LIBERO-Plus. Time-Domain Components implements temporal counterparts of the proposed components; each \emph{w/o} variant removes the indicated component. Standard CFG is the $\beta\rightarrow\infty$ limit of STB-CFG at the same $\lambda_0$.}
\label{tab:framework_ablation}
\end{table}

% \paragraph{Mechanism-level controls.}
% We provide three controlled comparisons in the supplementary material to separate the roles of frequency coordinates, training-scale adaptation, and inference guidance.
% The representation/source panel crosses temporal and DCT coordinates with white and spectrum-matched sources.
% The objective panel includes Raw MSE + Adaptive Weighting, which directly tests whether PSD normalization improves scale conditioning before adaptive weighting.
% The guidance panel evaluates no CFG, a fully swept CFG baseline, and STB-CFG using the same CFG-trained checkpoint.

\section{Conclusion}

We introduced FreqFM, a frequency-conditioned Flow Matching framework for VLA models.
By making action frequency explicit, FreqFM uses frequency-dependent statistics to shape the source distribution, adapt the training objective, and budget conditional guidance at inference.
The framework requires no VLA-backbone modification and improves over the paired temporal Flow Matching baseline by up to $9.3$ points on LIBERO, LIBERO-Plus, and VLA-Arena, and on six real-robot tasks.

\paragraph{Limitations.}
FreqFM relies on training-set-level spectral statistics: the spectrum-matched source is fixed across conditions, and the transport budget that limits the guidance residual is derived from the same fixed spectrum rather than from the current observation or language instruction.
The Extrapolation regression on VLA-Arena is the visible cost of this design, on the tasks whose action statistics are least likely to match the training spectrum.
FreqFM also requires the smooth action dimensions to be designated per embodiment.

\paragraph{Future work.}
Frequency statistics could instead be estimated conditioned on the current observation and instruction, which would address both the source distribution and the guidance budget from the same direction.
Future work could also test whether frequency-conditioned action generation transfers to other architectures, including world-action models.

\bibliography{references}

\end{document}